\documentclass[letterpaper, 10pt, conference]{ieeeconf}      %
\IEEEoverridecommandlockouts                              %

\title{
Behavior Tree Capabilities for Dynamic Multi-Robot Task Allocation
with Heterogeneous Robot Teams
}

\author{%
   Georg Heppner$^{1}$ \and David Oberacker$^{1}$ \and Arne Roennau$^{1}$ \and Rüdiger Dillmann$^{1}$%
  \thanks{$^{1}$\ Department of Interactive Diagnosis and Service
    Systems (IDS), FZI Research Center for Information Technology,
    Haid-und-Neu-Straße 10--14, 76131~Karlsruhe, Germany.
    }%
\thanks{
    The research leading to these results has received funding in the ROBDEKON II project under the grant agreement No. 13N16540 by the German Federal Ministry of Education and Research (BMBF).
}
}

\usepackage[usenames,dvipsnames,table,rgb]{xcolor}
\usepackage[pdfusetitle]{hyperref}
\usepackage[a-3u]{pdfx}
\usepackage{times}
\usepackage{amsmath}
\usepackage{amssymb}
\usepackage{amsthm}
\usepackage[squaren]{SIunits}
\usepackage{nicefrac}
\usepackage[utf8]{inputenc}
\usepackage{hyphenat}
\usepackage{xspace}
\usepackage{graphicx}
\usepackage{algorithm}
\usepackage{algorithmic}
\usepackage{xurl}
\usepackage{dsfont}
\usepackage{csquotes}
\usepackage{footmisc}
\usepackage{pifont}
\usepackage{multirow}
\usepackage{tabularx}
\usepackage{makecell}
\usepackage{adjustbox}

\usepackage{caption}
\usepackage{subcaption}

\usepackage{tikz}
\usetikzlibrary{decorations.pathreplacing, fit, calc, positioning, shapes,arrows,backgrounds}

\usepackage{pgfplots}
\DeclareUnicodeCharacter{2212}{−}
\usepgfplotslibrary{groupplots,dateplot}
\usetikzlibrary{patterns,shapes.arrows}
\pgfplotsset{compat=newest}

\usepackage[
    backend=biber,
    style=ieee,
    alldates=iso,
    seconds=true,
    sorting=ydnt,
    sortcites=true]{biblatex}

\hypersetup{
    hidelinks
}

\makeatletter%
\let\l@ngrel@x\protected\provide@command{\nbhyp}{%
  \nobreak\mbox{-}\nobreak\hskip\z@skip}
\makeatother%

\theoremstyle{definition}
\newtheorem{definition}{Definition}[section]

\usepackage[acronym,numberedsection=false]{glossaries}
\usepackage{glossary-mcols}
\glsdisablehyper
\makeglossaries

\newcommand{\rospkg}[1]{\texttt{#1}}

\newcommand{\mybtlib}{\rospkg{ros\_bt\_py}}

\newcommand{\rosmsg}[2][]{%
  \ifthenelse{ \equal {#1} {}}
  {}
  {\textcolor{black!75}{\rospkg{#1/}}\hspace{0pt}}%
  \texttt{#2}%
}

\newcommand*{\varname}[1]{\mathord{\mathit{#1}}}

\newcommand{\btgraph}{\mathcal{G}_T}
\newcommand{\btworld}{\mathcal{W}}

\newcommand{\btdataalph}{\Sigma}

\newcommand{\btparams}{\mathbb{P}}

\newcommand{\btcapabilities}{\mathcal{C}}
\newcommand{\btcapabilityimpl}{\mathcal{I}}
\newcommand{\btcapabilityrequirements}{\mathcal{R}}

\newcommand{\btnodes}{\mathcal{N}_T}
\newcommand{\btedges}{\mathcal{E}_T}
\newcommand{\btenv}{\mathit{Env}_T}
\newcommand{\btcapabilityenv}{\mathit{Env}^c_T}
\newcommand{\btcapabilityimplenv}{\mathit{Env}^i_T}
\newcommand{\btorderfunc}{\mathcal{o}}

\newcommand{\paramkind}{k}
\newcommand{\paramtype}{t}

\newcommand{\datagraph}{\mathcal{G}_D}
\newcommand{\datanodes}{\mathcal{N}_D}
\newcommand{\dataedges}{\mathcal{E}_D}

\newcommand{\crossmark}{\ding{55}}

\definecolor{fzi-green}{RGB}{13,114,73}

\definecolor{succeeded}{HTML}{1b5e20}
\definecolor{failed}{HTML}{b71c1c}
\definecolor{running}{HTML}{6A1B9A}%
\definecolor{idle}{RGB}{0,0,0}
\definecolor{uninitialized}{HTML}{455a64}
\definecolor{error}{HTML}{CC4700}
\definecolor{shutdown}{HTML}{1a237e}

\tikzset{
  treenode/.style = {draw=idle, align=center, inner sep=0pt, thick, text centered,
      font=\sffamily, text=black},
  btcondition/.style = {treenode, ellipse, inner sep=4pt, draw=black},
  btaction/.style = {treenode, rectangle, inner sep=4pt, draw=black},
  btcapability/.style = {btaction, very thick, dashed},
  btdecorator/.style = {treenode, chamfered rectangle, inner sep=4pt, draw=black},
  btnode/.style = {treenode, rectangle, inner sep=4pt, black, draw=black, text height=0.8em, text width=1.2em},
  round/.style={rounded corners=1.5mm,minimum width=1cm,inner sep=2mm,above right,draw},
  bt-idle/.style={draw=idle, thick},
  bt-uninitialized/.style={draw=uninitialized, thick},
  bt-succeeded/.style={draw=succeeded, thick},
  bt-failed/.style={draw=failed, thick},
  bt-running/.style={draw=running, thick},
  bt-error/.style={draw=error, thick},
  bt-shutdown/.style={draw=shutdown, thick}
}

\tikzset{fontscale/.style = {font=\relsize{#1}}}

\newacronym{ros}{ROS}{Robot Operating System}
\newacronym[longplural={Behavior Trees}]{bt}{BT}{Behavior Tree}
\newacronym[longplural={Forschungszentrum Informatik}]{fzi}{FZI}{FZI Research Center for Information Technology}
\newacronym[longplural={Remote Procedure Calls}]{rpc}{RPC}{Remote Procedure Call}
\newacronym[longplural={Domain Specific Languages}]{dsl}{DSL}{Domain Specific Language}
\newacronym[longplural={Hierachical Finite State Machines}]{hfsm}{HFSM}{Hierachical Finite State Machine}
\newacronym{asymtre}{ASyMTRe}{Automated Synthesis of Multi-robot Task solutions through software}
\newacronym{ctas}{CTAS}{Capability-based robust Task Assignment and Scheduling}
\newacronym{cnp}{CNP}{Contract Net Protocol}
\newacronym{mrtap}{MRTAP}{Multi-Robot Task Allocation Problem}
\newacronym{slam}{SLAM}{Simultaneous localization and mapping}
\newacronym{comutar}{CoMutaR}{Coalition formation based on Multitasking Robots}

\newcommand\copyrighttext{%
	\footnotesize \textcopyright \the\year{} IEEE. Personal use of this material is permitted.
	Permission from IEEE must be obtained for all other uses, in any current or future
	media, including reprinting/republishing this material for advertising or promotional
	purposes, creating new collective works, for resale or redistribution to servers or
	lists, or reuse of any copyrighted component of this work in other works.}
\newcommand\copyrightnotice{%
	\begin{tikzpicture}[remember picture,overlay]
		\node[anchor=south,yshift=10pt] at (current page.south) {\fbox{\parbox{\dimexpr\textwidth-\fboxsep-\fboxrule\relax}{\copyrighttext}}};
	\end{tikzpicture}%
}

\begin{document}

\maketitle
\thispagestyle{empty}
\pagestyle{empty}

\begin{abstract}
    While individual robots are becoming increasingly capable, with new sensors and actuators, the complexity of expected missions increased exponentially in comparison.
    To cope with this complexity, heterogeneous teams of robots have become a significant research interest in recent years.
    Making effective use of the robots and their unique skills in a team is challenging.
    Dynamic runtime conditions often make static task allocations infeasible, therefore requiring a dynamic, capability-aware allocation of tasks to team members.
    To this end, we propose and implement a system that allows a user to specify missions using \glspl{bt}, which can then, at runtime, be dynamically allocated to the current robot team.
    The system allows to statically model an individual robot's capabilities within our \textit{ros\_bt\_py} \gls{bt} framework.
    It offers a runtime auction system to dynamically allocate tasks to the most capable robot in the current team.
    The system leverages utility values and pre-conditions to ensure that the allocation improves the overall mission execution quality while preventing faulty assignments.
    To evaluate the system, we simulated a find-and-decontaminate mission with a team of three heterogeneous robots and analyzed the utilization and overall mission times as metrics.
    Our results show that our system can improve the overall effectiveness of a team while allowing for intuitive mission specification and flexibility in the team composition.
\end{abstract}

\copyrightnotice

\section{Introduction}

With the widespread adoption of increasingly complex robotic systems, the need for new methods of defining a system's high-level behavior while simultaneously considering complex skill sets is becoming apparent.
Not only are specialized systems, such as walking robots or AI-assisted drones, commercially available from multiple manufacturers, but they are becoming more versatile with added manipulators or specialized 3D sensors, making teams of heterogeneous robots a very realistic proposition.
However, coordinating very heterogeneous robots and skills while dealing with dynamic team compositions requires a robust and flexible system to model, coordinate, and exchange robot skills and behaviors.
Therefore, implementing a decentralized system optimized to solve the \emph{\gls{mrtap}} for heterogeneous robots is a worthwhile endeavor.
Solving the \gls{mrtap} has been a topic in robotics for many years.
\citeauthor{gerkey_formal_2004}~\cite{gerkey_formal_2004} provide a classification of sub-problems to the \gls{mrtap}.
In the following we will look primarily at the \enquote{\textbf{S}ingle \textbf{T}ask - \textbf{M}ulti \textbf{R}obot - \textbf{I}nstantanious \textbf{A}ssignment (\textit{ST-MR-IA})} sub-problem.%
\begin{figure}[ht]
    \centering
    \includegraphics[width=0.9\columnwidth]{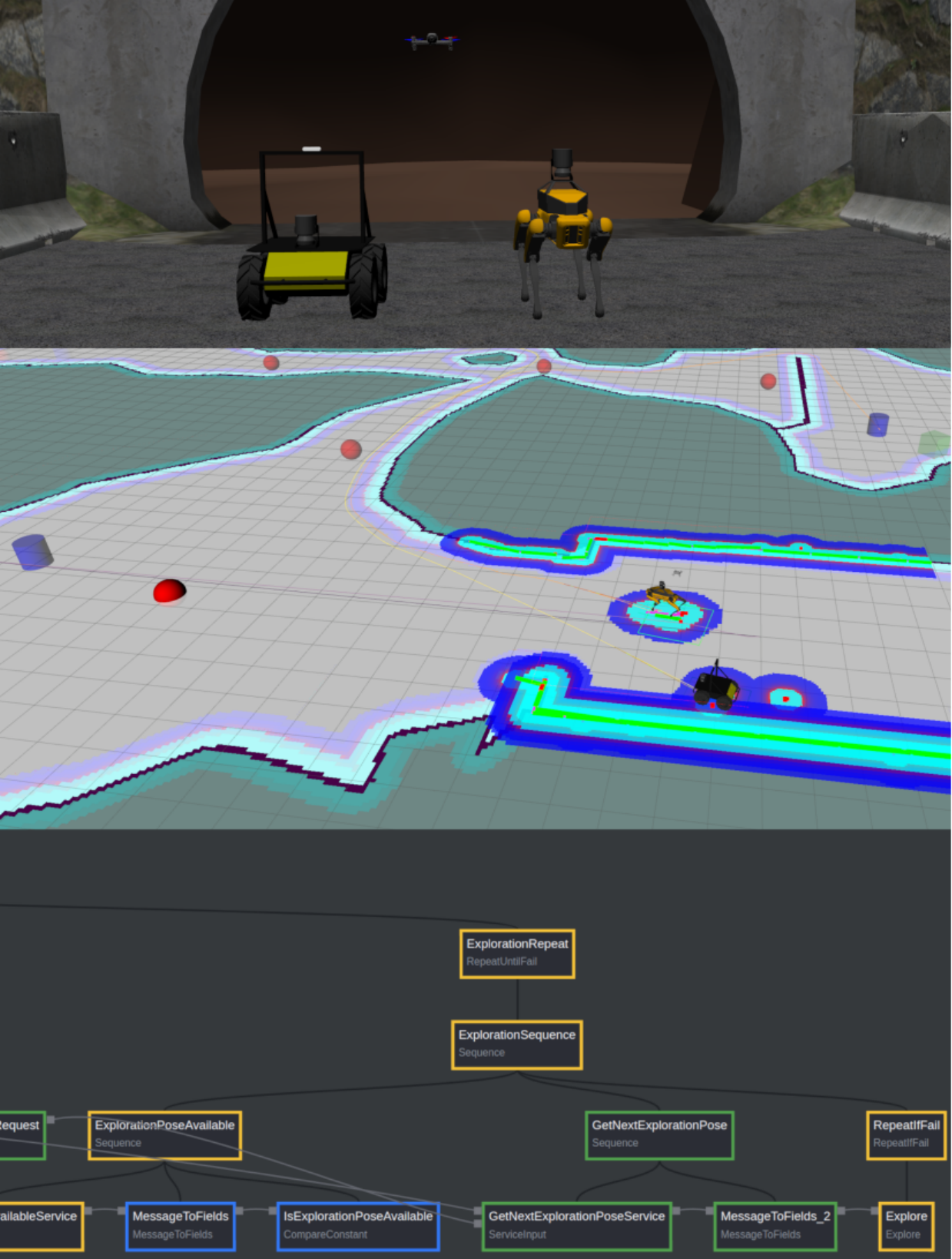}
    \caption[Simulation Environment, RVIZ visualization and Part of the Behavior Tree for the simulated mission]{
        Top: Gazebo Simulation with a heterogeneous robot team, Middle: 2D Cave map with target objects (blue) exploration goals (red) and paths. All three robots employ this map for navigation with the ROS \texttt{move\_base} stack, Bottom: Behavior Tree segment used in the Explore Capability.
    }
    \vspace{-15px}
    \label{fig:simulation}
    \label{fig:cave_world:navigation}
\end{figure}

\par
\acrlongpl{bt} have emerged as modular and easy-to-use method for specifying robot behavior~\cite{IOVINO2022104096, 10.1145/3426425.3426942}.
Works by~\citeauthor{ogren_increasing_2012}~\cite{ogren_increasing_2012} and \citeauthor{marzinotto_towards_2014}~\cite{marzinotto_towards_2014,Colledanchise17,colledanchise_implementation_2021,colledanchise_behavior_2018,TRO17Colledanchise} propose the foundational concepts for~\glspl{bt} in robotics.
Early on, they were also proposed for multi-robot systems.
\cite{colledanchise_advantages_2016} outlines how \glspl{bt} can be utilized for controlling multi-robot setups.
\citeauthor{10.1109/AIM52237.2022.9863364}~\cite{10.1109/AIM52237.2022.9863364} describe a centralized method for controlling multiple mobile robots using a singular \gls{bt}.
To a similar end, we also proposed a~\gls{bt} framework for controlling teams of heterogeneous robots called \texttt{ros\_bt\_py}~\cite{heppner_ros_bt_py_2023}.
While the previously mentioned approaches allow the remote execution of static, predefined sub-trees, no approach can handle heterogeneous or dynamic abilities in the robot team yet.
Non Behavior Tree approaches to the \gls{mrtap} have focused more on methods for modeling robot capabilities as a core challenge.
The ASyMTRe framework proposed in~\cite{tang_asymtre_2005,zhang_iq-asymtre_2013} facilitates collaboration by automatically synthesizing a data flow graph between sensors, processors, and actuators for a given team and set of tasks.
In the CoMutaR framework~\cite{shiroma_comutar_2009} this principle is extended to account for resource limitations and dynamic changes during runtime, at the cost of a sizable modeling overhead.
\citeauthor{guerin_framework_2015}~\cite{guerin_framework_2015} propose a robot-independent mission \gls{bt} format in addition to a model for a robot's abilities.
The approach is focused on single robot industrial applications where tasks should be described independently of the robot executing them.
Assigning tasks to teams with degrading abilities over time is covered by~\cite{irfan_auction-based_2016}.
While the team composition is dynamic for this approach, it requires a static set of available capabilities.
A similar problem is solved in~\cite{fu_robust_2022}, where capabilities and task suitability are modeled as vectors, allowing for a probabilistic approach to determine an optimal task allocation over the mission runtime.
\par
The \gls{mrtap} has been explored more thoroughly for homogeneous robot teams.
Older approaches like \textit{MURDOCH}~\cite{gerkey_sold_2002} and \textit{TraderBots}~\cite{zlot_multi-robot_2002, dias_robust_2004, dias_traderbots_2004} aim to group related tasks and allocate them using auctions.
More recent approaches include \cite{rajchandar_novel_2022,alshaboti_multi-robot_2021,wen_multi-robot_2021,eijyne_development_2020,fu_robust_2022}, which focus on task synergies and optimizing multiple metrics during allocation.
Current approaches for \gls{mrtap} are capable of generating optimized allocations for static robot teams, often using a centralized authority.
Yet task allocations for dynamic teams of robots with heterogeneous capabilities still have many challenges that need to be addressed.
Similarly, the restriction to a centralized task allocation process introduces problems for mobile robots and spontaneous cooperation.
Thus, our goal is to realize a decentralized multi-robot task allocation system specifically for teams of heterogeneous robots.
This includes providing an abstraction for the heterogeneous skills of individual team members.
Decentralization is realized by employing an auction-based approach that enables each robot to act as the auctioneer, voiding the need for a centralized authority.
These approaches are realized as an extension to the \textit{ros\_bt\_py} \acrfull{bt} framework by introducing \textit{capabilities} as a skill abstraction that can dynamically be auctioned off to robots in the same team.

In section \ref{sec:approach}, we present the theoretical model of our proposed capabilities.
The implementation of the model as part of the open-source \textit{ros\_bt\_py} library is described in section \ref{sec:implementation}.
Section \ref{sec:evaluation} details our experimental evaluation of the system using a simulated find-and-decontaminate mission.
Finally, section \ref{sec:conclusions} summarizes this work and provides an outlook for future works.

\section{Capabilities in Behavior Trees for MRTA}
\label{sec:approach}
Our previously proposed \acrlong{bt} framework~\cite{heppner_ros_bt_py_2023} implements common \gls{bt} functionality such as flow control, ticking, and sub-trees to model individual capabilities but can also be used to design complete missions by enabling long running tasks and utility evaluation of sub-trees.
Finally, the possibility to execute parts of the \gls{bt} on a distant system enables multi-robot task allocation for a team of heterogeneous robots as the mission can be spread over a group of robots based on their individual skills.
However, the proposed method relies on \emph{shovables}, a special decorator node that explicitly marks sub-trees suitable for remote execution, which have to be evaluated at the startup of the mission tree.
This limits the possibility of changing the assignment during runtime, restricts reactivity of the parameters and results, and limits the complexity of the sub-trees.
We therefore extended our previous work by specifically adding \emph{capabilities} to the model and enable a dynamic distribution via \emph{auction} of these within the team.
All following definitions build upon the concepts \& definitions introduced in \cite{heppner_ros_bt_py_2023}.

\subsection{Capabilities in Behavior Trees}
\label{subsec:approach:capabilities}
\emph{Capabilities} abstract individual robot skills as node representations of semantically meaningful operations, e.g., \texttt{MoveToPose} or \texttt{OpenDoor}, in a \gls{bt}.
They do this by providing functionality similar to a sub-tree but are assigned at runtime to enable a dynamic evaluation, usage of a suitable implementation, distribution of its execution via auction, and handling input and output parameters across systems.
Before ticking the \gls{bt}, the actual sequence of \gls{bt} nodes that perform the skill described by the capability is not known, which requires some special handling.

\begin{definition}[Capabilities]
    \label{def:approach:capabilities}
    A capability $c$ is a Node in the \gls{bt} that provides an abstract representation of a composite action with a single semantic meaning.
    \begin{equation*}
        \forall c \in \btcapabilities: c \in \btnodes \\
    \end{equation*}
    Where $\btnodes$\ denotes the set of all nodes in a \gls{bt} and $\btcapabilities$ the set of all capabilities.
    Capabilities are executor leaf nodes.
\end{definition}

\begin{definition}[Parameters for Capabilities]
    A single Parameter $p$ is given with the Triple $p=(n, k, t)$ consisting of a node $n$, kind $k$ (input, output, option) and type $t$.
    $\btparams$ denote the set of all parameters, $\mathcal{P}(\btparams{})$ its power set.
    \begin{equation*}
        \begin{split}
            \varname{inputs}&: \btcapabilities \to \mathcal{P}(\btparams{}) \qquad \varname{outputs}: \btcapabilities \to \mathcal{P}(\btparams{}) \\
            \varname{inputs}&(c) =  \{ p = (n, \paramkind{}, \paramtype{}) \mid  p \in \btparams{} \land n = c \land \paramkind{} = \text{input} \}\ \\
            \varname{outputs}&(c)  =  \{ p = (n, \paramkind{}, \paramtype{}) \mid  p \in \btparams{} \land n = c \land \paramkind{} = \text{output} \}
        \end{split}
    \end{equation*}
    Capabilities cannot have option values, as they are incompatible with the dynamic binding:
    \begin{equation*}
        \forall c \in \btcapabilities: \neg \exists p \in \btparams{} : p=(n,\paramkind{}, \paramtype{}) \mid n = c \land k = \text{option} \\
    \end{equation*}
\end{definition}

\begin{definition}[Capability Implementation]
    \label{def:capability_impl}
    A capability node only provides the model information of the functionality.
    To actually do something a capability implementation $i$ is required.
    An implementation $i$ contains a complete \gls{bt}, given as BT-graph $\btgraph{}(\btnodes{}, \btedges{}, \btorderfunc{})$ where $\btnodes{}$ contains all of the nodes which might also be capabilities, $\btedges{}$ defines the edges between them, $\btorderfunc{}$ their order.
    Each implementation is specific to a single robot, specified by its world state $\btworld \in \Sigma^{x}; x \in \mathbb{N}$.
    The world state is build with an arbitrary data alphabet $\Sigma$ which models the know information about the world for a single robot, including its sensor data and other states.
    The data-graph $\datagraph{}(\datanodes{}, \dataedges{})$ models the data interactions with the parameters where $\datanodes{}$ are the data nodes and $\dataedges{}$ their edges.
    \begin{equation*}
        i = (\btworld, \btgraph{}(\btnodes{}, \btedges{}, \btorderfunc{}), \datagraph{}(\datanodes{}, \dataedges{}))
    \end{equation*}
\end{definition}

\begin{definition}[Capability to Implementation connection]
    \label{def:capability_to_impl}
    Robots can exchange parts of their world state with the function $\varname{nearbyWorlds}(\btworld)$, therefore  $\btcapabilityimpl$ denotes the set of all capability implementations over all robots currently in the same team.
    By explicitly marking a capability as local, only locally available implementations are considered.
    Each implementation~$i$ corresponds to a single capability $c$.
    The association between the two is defined by the $\varname{ c_{impl}}$ function:
    \begin{equation*}
        \varname{c_{impl}} : \btcapabilityimpl \to \btcapabilities \\
    \end{equation*}
    $\varname{ c_{impl}}$ is surjective but not injective, meaning every capability is associated with one up to infinity implementations.
    The $\varname{impl}$ function on the other hand returns all currently executable implementations for a given capability $c$:
    \begin{equation*}
        \begin{split}
            impl: \btcapabilities \to & \mathcal{P}(\btcapabilityimpl) \\
            \varname{impl}(c) =& \{i \mid i \in I \land \varname{c_{impl}}(i) = c \land \\
            &\quad \varname{executable}(c, i) \land \varname{validate_{precondition}}(i) \}
            \}
        \end{split}
        \label{eq:impl_func}
    \end{equation*}
\end{definition}

\begin{definition}[Capability Execution]
    \label{def:capability_execution}
    An implementation can be executed if it implements a capability, fulfills all the preconditions and has a valid execution location.
    \begin{equation*}
        \begin{split}
            \varname{executable}(c, i) = & (\btworld_i = \btworld_c) \lor                                                                                                  \\
            & (\lnot~\varname{{required}_{local}}(c) \land                                                                   \\
            & \exists \btenv = (\btgraph(\btnodes{}, \btedges{}, \btorderfunc{}), \_, \_ , \btworld_i): \\
            & \exists n \in \btnodes{} : n = \text{RemoteCapabilitySlot})
        \end{split}
    \end{equation*}
    $\btworld_i$  denotes the world state where the implementation $i$ is located, $\btworld_c$ the world state where the capability $c$ is located.
    This is either the case if the implementation is on the same robot as the interface or remote execution is allowed and the robot that has the implementation currently has a \textit{RemoteCapabilitySlot}.

\end{definition}

\begin{definition}[Preconditions]
    \label{def:concept:capabilities:preconditions}
    To ensure that a given capability implementation $i$ can be executed correctly during runtime preconditions can be specified.
    Preconditions can require a set of capability interfaces to be executed successfully before the implementation runs.
    For each required capability it can be specified if they should have been executed on the local robot or on any nearby robot.
    $\btcapabilityrequirements$ describes the set of all possible preconditions.
    \begin{equation*}
        \btcapabilityrequirements=\{(c, k) \mid c \in \btcapabilities; k \in \{\text{local}, \text{remote}\}\}
    \end{equation*}
    The precondition function returns all preconditions for a given capability implementation.
    \begin{equation*}
        \text{precondition}: \btcapabilityimpl \to~\mathcal{P}(\btcapabilityrequirements)
    \end{equation*}
    Before $i$ is executed every precondition is validated against the current runtime context:
    \begin{equation*}
        \begin{split}
            \varname{{validate}_\text{precondition}}(i) &= \forall r =(c, k) \in \varname{precondition}(i): \\
            &\left( k = \text{local} \land c \in \btnodes \right) \lor \\
            &\left( k = \text{remote} \land \exists \btgraph(\btnodes', \btedges', \btorderfunc') \right. \\
            &\left. \qquad \in \varname{nearbyTrees}(\btworld): c \in \btnodes' \right)
        \end{split}
    \end{equation*}
    Where  $\varname{nearbyTrees}$ is a function that returns the \gls{bt} of nearby worlds.
    If a precondition cannot be validated the implementation execution will fail before ticking any node.
\end{definition}

\begin{figure}[t]
    \centering
    \includegraphics[width=\columnwidth]{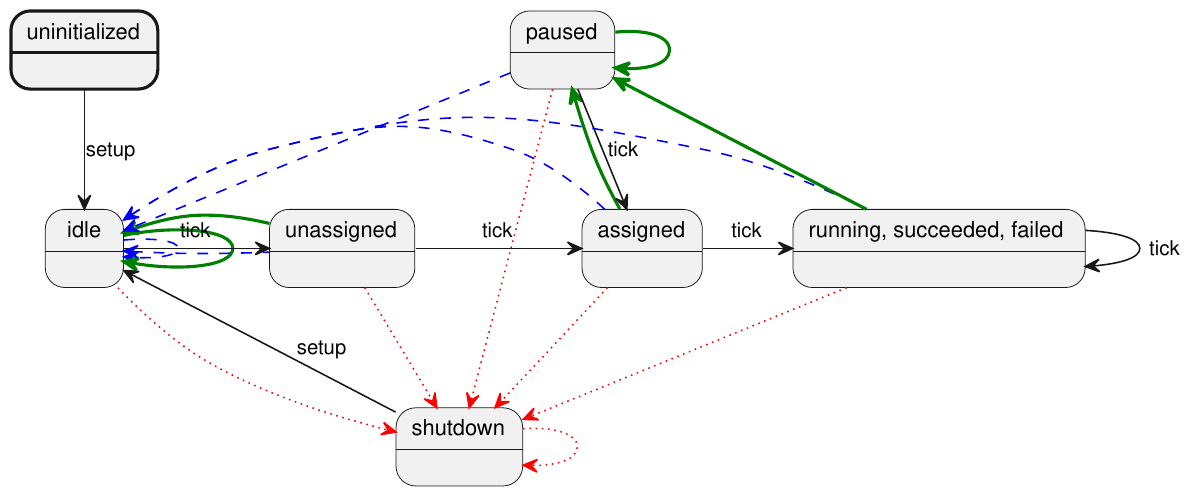}
    \caption[State transitions for \emph{Capability} \gls{bt} nodes]{
        State transitions for \emph{Capability} \gls{bt} nodes when applying actions.
        For clarity the error state/action has been left out.
        To further simplify: the red dotted edges are representing the \texttt{shutdown} action, blue dashed edges the \texttt{reset} action and the green solid lines the \texttt{untick} action.
        Lastly, the running, succeeded, and failed state haven been combined.
    }
    \label{fig:concept:capabilities_bt:c_states}
    \vspace{-10px}
\end{figure}
\begin{definition}[Dynamic Binding]
    \label{def:concept:capabilities:dynamic_binding}
    As the capabilities are evaluated at runtime (dynamic binding), the node State definition (\cite{heppner_ros_bt_py_2023})     needs to be extended to handle changing implementations during runtime.
    Two additional states are added:
    \begin{equation*}
        \begin{split}
            \varname{state}(n) \in \btnodes = \{&uninitialized, error, idle,\\
            &succeeded, failed, running,\\
            &shutdown, \textbf{unassigned, assigned}\}
        \end{split}
    \end{equation*}
    As the states can apply to all nodes, for example if a control flow node contains a capability, the behavior needs to be defined for all of them when using the update function $\varname{update}(n, a, {Env}_{T})$ which determines the new node world state after applying an action $a$.
    For control-flow nodes, the new states act identical to the running state.
    Non-capability leaf nodes cannot reach the new states, thus they are not relevant for them.
    Capabilities follow the state transitions outlined in figure~\ref{fig:concept:capabilities_bt:c_states}.
    An assignment function $\varname{assign_c}(c)$ allows retrieving the currently assigned implementation.
    \begin{equation*}
        \begin{split}
            &\varname{assign_c}(c) : \btcapabilities \to \btcapabilityimpl \cup \{\varnothing\} \\
            &c \to
            \begin{cases}
                \varnothing             & \varname{state}(c) \in \{idle, shutdown, error, \\ & \qquad \qquad \quad unassigned\} \\
                i \in \varname{impl}(c) & \varname{state}(c) \in \{assigned, succeeded,   \\
                                        & \qquad \qquad \quad  running, paused, failed\}
            \end{cases}
        \end{split}
    \end{equation*}
    When returning an implementation the value out of the set $\varname{impl}(c)$ is determined in the \emph{unassigned} state.
    While in the \textit{assigned, paused, running, succeeded, failed} state, the value returned by $\varname{impl}(c)$ has to remain unchanged.
    Upon receiving a reset or shutdown action, the value will revert to $\varnothing$.
\end{definition}

\subsection{Dynamic Capability Execution}
The \emph{shovables} we proposed in our previous work enable the remote execution of a sub-tree by extracting relevant parts of the $\btenv$ and forwarding it to a \emph{slot} selected during setup of the tree where the execution is handled before an updated environment $\btenv'$ is returned and integrated.
Changes of the parameters during execution are not considered, and the \emph{slot} can not be changed, both requirements to enable dynamic capability execution.
To tackle this, and to ensure that updates are passed between the Environment $\btenv$ of the capability and the environment $\btcapabilityimplenv$ of the implementation, which do not need to be the same, \emph{CapabilityIOBridges} are introduced:

\begin{definition}[Capability IO Bridge]
    For any given capability $c$ there exists  an $c_{\text{in}}$ (Input Bridge) and an $c_{\text{out}}$ (Output Bridge):
    \begin{align*}
        \forall c \in \btcapabilities: \exists c_{\text{in}} \in \btcapabilities_{in}:   & ~c_{\text{in}} =~\varname{br_{in}}(c)   \\
        \forall c \in \btcapabilities: \exists c_{\text{out}} \in \btcapabilities_{out}: & ~c_{\text{out}} =~\varname{br_{out}}(c)
    \end{align*}
    $\varname{br_{in}}(c)$ and $\varname{br_{out}}(c)$ are bijective functions associating bridges with capabilities.
    $\tilde{\varname{br_{in}}}(c_{br})$  and $\tilde{\varname{br_{out}}}(c_{br})$  are the respective inverse functions.
    Similarly to capabilities, both bridges are~\gls{bt} leaf nodes.
\end{definition}

\begin{definition}[Capability Input and Output Bridge]
    Capability input and output bridges pass the parameters from the capability to the implementation and back by acting as adapter.
    The input bridge nodes publishes the inputs passed to the capability as outputs, while the output bridge node similarly provides inputs that are passed to the outputs of the capability:
    \begin{equation*}
        \begin{split}
            \varname{outputs}(c_{in}) : \btcapabilities_{in} \to& \mathcal{P}(\btparams) \\
            c \to&  \{ p = (n, k, t) \mid  p \in \varname{inputs}(\tilde{\varname{br_{in}}}(c_{in})) \}\\
            \varname{inputs}(c_{out}) : \btcapabilities_{out} \to& \mathcal{P}(\btparams) \\
            c \to&  \{ p = (n, k, t) \mid  p \in \varname{outputs}(\tilde{\varname{br_{out}}}(c_{in})) \}
        \end{split}
    \end{equation*}

    Input bridges do not have inputs, while output bridges do not have outputs:
    \begin{equation*}
        \begin{split}
            \varname{inputs}(c_{in}) : \btcapabilities_{in} \to& \mathcal{P}(\btparams) \\
            \varname{outputs}(c_{out}) : \btcapabilities_{out} \to& \mathcal{P}(\btparams) \\
            c \to& ~\emptyset
        \end{split}
    \end{equation*}

    While the state of the capability interface $\varname{state}(c) = \text{running}$, during each call of $\btenv' = update(c, \text{tick}, \btenv)$, the following is ensured after every tick:
    \begin{equation*}
        \begin{aligned}
            \forall c \in \btcapabilities : & c_{\text{in}} = \varname{br_{in}}(c):                                          \\
                                            & \forall p = (n, k, t) \in \varname{inputs}(c) :                                \\
                                            & \quad \forall p' = (n', k', t') \in \varname{outputs}(c_{in}):                 \\
                                            & \qquad n = n'                                                                  \\
                                            & \qquad \implies \varname{value}(p, \btworld) = \varname{value}(p', \btworld'), \\
            \forall c \in \btcapabilities : & c_{\text{out}} = \text{br}_{\text{out}}(c):                                    \\
                                            & \forall p = (n, k, t) \in \varname{outputs}(c) :                               \\
                                            & \quad \forall p' = (n', k', t') \in \varname{inputs}(c_{out}):                 \\
                                            & \qquad n = n'                                                                  \\
                                            & \qquad \implies \varname{value}(p, \btworld) = \varname{value}(p', \btworld').
        \end{aligned}
    \end{equation*}
    Where $\varname{value}(p, W)$ is the parameter value function.
    $W$ and $W'$ are the respective world states of the capability interface and the implementation.
\end{definition}

In order to execute the capabilities on arbitrary system the \emph{RemoteCapabilitySlot} is introduced:

\begin{definition}[Remote Capability Slot]
    \label{def:remote_capability_slot}
    A remote capability slot $c_r$ is a Node that can execute local capability implementations $i \in \btcapabilityimpl$ on request from remote robots.
    Only one implementation can be executed at a time.
    Upon receiving an execution request, containing the capability $c$ and the requested implementation $i = (\btworld, \btgraph{}(\btnodes{}, \btedges{}, \btorderfunc{}), \datagraph{}(\datanodes{}, \dataedges{}))$, the node creates a new execution environment called $Env^{c}_{T}$ with:
    \begin{equation*}
        Env^{c}_{T} = (\btgraph{}, \datagraph{}, \btdataalph{}, \btworld{})
    \end{equation*}
    This execution environment exists in the world state of the~\gls{bt} $c_r$ is a part of, but uses an independent $\btgraph{}$ and $\datagraph{}$.
    Actions for the new environment $Env^{c}_{T}$ are passed through the $c_r$ node, similar to a decorator node.
    Thus, the $Env^{c}_{T}$ is ticked at the same rate as the environment of $c_r$.
    Parameter values within the  $Env^{c}_{T}$ are exchanged with the remote robots $c$ by using the \emph{CapabilityIOBridges} and the state of the root of $\btgraph$ is transmitted.
    When an implementation finishes its execution, all traces of it are removed from the environment the \emph{RemoteCapabilitySlot} is part of, resetting it to the state before accepting the request.
    It uses a similar state machine as regular capability interfaces, depicted in figure~\ref{fig:concept:capabilities_bt:c_states}.
    Main differences are that the unassigned, assigned and running state all revert back to the unassigned state after unticking and that the transition form unassigned to the assigned state requires an external signal originating from the capability that has requested the remote operation.
\end{definition}

\subsection{Capability Behavior}

To clarify the behavior of the capability, their respective behaviors during the relevant states are explained:

\begin{definition}[Behavior in the Unassigned State]
    \label{def:concept:capabilities:unassigned}
    When $\btenv' = update(c, \text{tick}, \btenv)$ is called while $\varname{state}(c) = \textit{unassigned}$, the capability will try to determine a concrete return value for the $\varname{assign_c}(c)$ function.
    The assignments will be determined by an auction-based system.
    In ${Env'}_{T}$ the following holds: $\varname{state}(c)=\textit{assigned}$ and $\text{assign}_c(c) \neq \varnothing$.
\end{definition}
\begin{definition}[Behavior in the Assigned State]
    \label{def:concept:capabilities:assigned}
    When $\btenv' = update(c, \text{tick}, \btenv)$ is called while $\varname{state}(c) = \textit{assigned}$, the capability interface will setup the tree specified by the $\varname{assign_c}(c)$ function.
    When a implementation on the local robot is selected for execution, it will be setup similarly to the sub-trees specified in~\cite{heppner_ros_bt_py_2023} with the slot being the capability $c$ itself.
    If the execution is on a remote robot, the implementation will be setup in an available \textit{RemoteCapabilitySlot}.
    The created execution environment, either local or remote, is called~$Env^{c}_{T}$, with $r \in \btnodes$ representing the root of the implementation~\gls{bt}.
    If this completes, the node will proceed to the running state.
\end{definition}
\begin{definition}[Behavior in the Running State]
    \label{def:concept:capabilities:running}
    When  $\btenv' = update(c, \text{tick}, \btenv)$  is called while $\varname{state}(c) = \textit{running}$, the capability interface will forward the tick to the implementation execution environment.
    As described later, before the tick, the current input values of $inputs(c)$ will be passed into~$Env^{c}_{T}$ using the \emph{CapabilityIOBridge}.
    ${\btcapabilityenv}^{'} = update(r, \text{tick}, \btcapabilityenv)$ is called.
    The new values for $outputs(c)$ are extracted from  ${\btcapabilityenv}^{'}$ , using the \textit{CapabilityIOBridge}, as well as the state $state(r) \in S$.
    $\varname{state}(c) = \varname{state}(r)$ is set and the values for $outputs(c)$ are updated in $\btenv'$.
\end{definition}

\section{Implementation}
\label{sec:implementation}
\begin{figure}[t]
    \centering
    \includegraphics[width=\columnwidth]{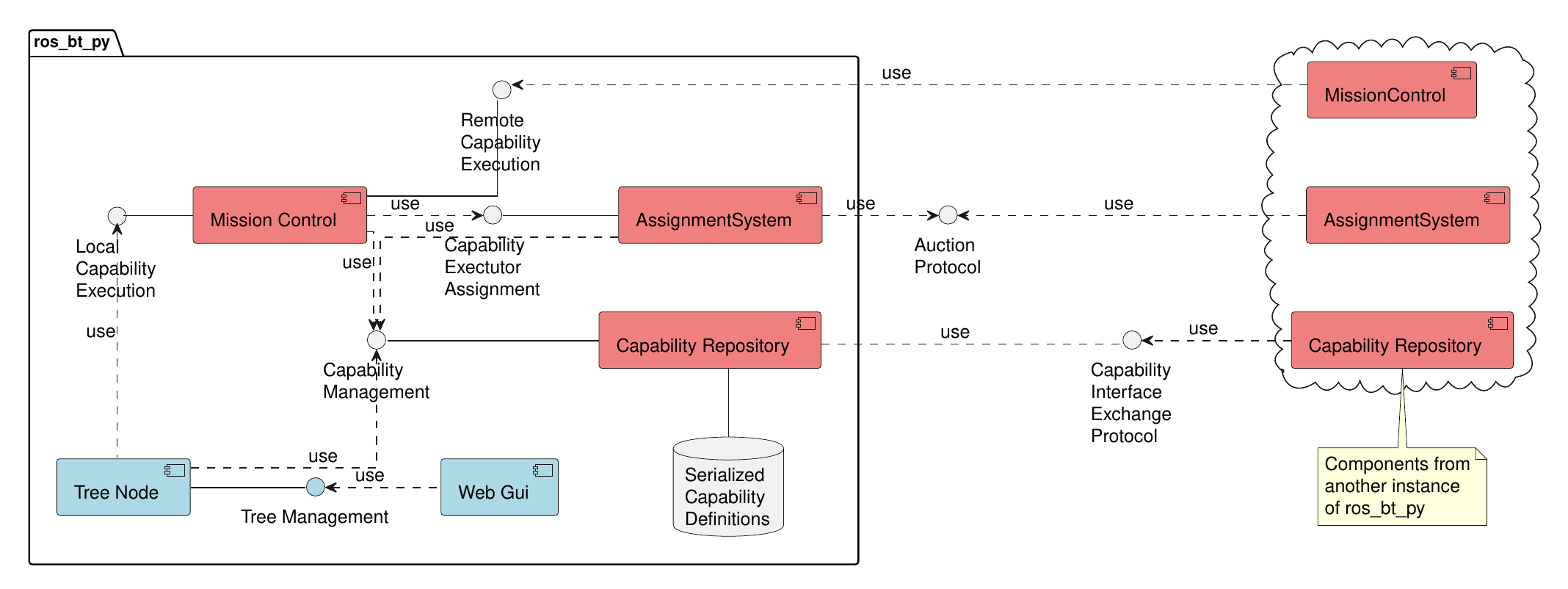}
    \caption[Schematic overview of the proposed architecture]{
        Schematic overview of the components required to implement the concept.
        Components marked in red, were implemented as part of this work.
    }
    \label{fig:concept:architecture}
    \vspace{-10px}
\end{figure}
The presented concept was implemented as an extension to our \texttt{ros\_bt\_py} framework~\cite{heppner_ros_bt_py_2023} (Figure~\ref{fig:concept:architecture}).
Each component is implemented as a \gls{ros} node written in Python, whereas the interfaces are \gls{ros} services or \gls{ros} topics.
\par
\emph{Capabilities}, \emph{RemoteCapabilitySlots}, \emph{IO inputs} and \emph{IO outputs} are implemented as abstract \gls{bt} nodes, which can be instantiated for specific input and output configurations.
\emph{Capability Implementations} are realized as sub-trees, where only IO input and output nodes are used to pass information across the sub-tree boundaries.
\par
The \textbf{Capability Repository} implements a distributed database of all available capabilities within the current team.
Additionally, it manages the available capability implementations on the local robot.
Both capabilities and implementations can be added and removed during runtime.
\par
Local and remote execution management for capabilities is realized in the \textbf{Mission Control} component.
For local execution, it selects the best local implementation from the capability repository based on calculated utility values~\cite{heppner_ros_bt_py_2023}.
To realize remote execution, the best local implementation, and the corresponding capability are passed to the \textbf{Assignment System}.
Upon receiving the best available remote robot, the local mission control will trigger the execution by communicating with the remote mission control.
During local or remote execution, it will receive heartbeat messages with reports on execution progress.
\par
The aforementioned \textbf{Assignment System} is an abstract component for determining the best available team member to execute a capability.
In the reference implementation, the component is realized as an open single-item auction system with re-auctions during execution time.
During the bid period, the weighted utility values from all eligible robots in the team are collected.
Robots are greedy bidders running as many tasks as possible while minimizing costs.
\par
The implementation of \texttt{ros\_bt\_py} was made public as open source project \footnote{\url{https://github.com/fzi-forschungszentrum-informatik/ros_bt_py}} and has recently been updated with the presented capability support.
\section{Evaluation}
\label{sec:evaluation}

\begin{figure}[t]
    \centering
    {\sffamily
        \begin{tikzpicture}[->,>=stealth',level/.style={sibling distance = 4.75cm/#1, level distance = 1.1cm}]
            \node [btnode] (parallel) {$||$}
            child { node [btcapability] (explore) {Explore}}
            child { node [btnode] (sel) {$\rightarrow$}
                    child { node [btcapability] (identify) {Identify \\ Object}}
                    child { node [btcapability] (movetodecon) {Move to \\ Decontamination}}
                }
            ;
            \node[round,draw=fzi-green,label={[fzi-green]above left=:Decontamination}, fit=(identify)(movetodecon)] {};
            \node[round,draw=fzi-green,label={[fzi-green]:Exploration}, fit=(explore)] {};
            \draw [-, color=black!75] (identify) edge [->, out=0, in=180] (movetodecon);
        \end{tikzpicture}
    }
    \caption{
        Simplified Behavior Tree describing the overall mission.
        Each Action node is a \emph{Capability} that is assigned to a robot for execution.
        The root node is a \textbf{parallel} node that ticks all its children in parallel.
    }
    \vspace{-15px}
    \label{fig:mission_bt}
\end{figure}
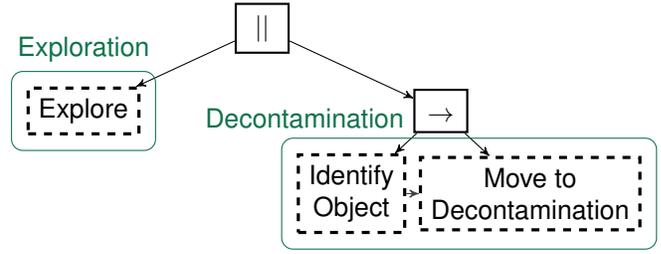

To evaluate our previously presented approach, we simulate a heterogeneous robot team performing a find-and-decontaminate mission in a cave environment.
The goal is for the system to automatically assign the different tasks within the mission to capable robots while adapting the plan to current runtime conditions.
Task assignment is based on the position of the robot and the robots ability to perform the task.
Robot failures, team expansions, and uncertainty of successful task completion were induced to evaluate the system's ability to adapt during runtime.

\subsection{Mission}
\label{sec:evaluation:mission}
The primary mission is to find contaminated objects in the cave and move them to the decontamination area.
Thus, there are three tasks for the robots to execute:
\textbf{Explore} reveals nearby objects and further exploration points, \textbf{Identify} checks if a found object is contaminated, and \textbf{Decontaminate} moves contaminated objects to the decontamination area.
As the aim of the evaluation is the task allocation, we simplified the task execution by using a \emph{simulation manager}.
The simulation manager offers \gls{ros} services to simulate the sensor results and manipulation operations during task execution.
To make the simulated execution more realistic, the simulation manager waits for randomized intervals before returning results, with a randomized chance for each operation to fail.
In figure~\ref{fig:cave_world:navigation}, the tasks 2D-Poses within the map can be seen.
Figure~\ref{fig:mission_bt} shows the simplified \gls{bt} encoding the overall mission behavior.
Found object poses during exploration are exchanged using a globally shared database.
\subsection{Simulation Environment}
\label{sec:evaluation:sim}
Gazebo\footnote{\url{http://wiki.ros.org/gazebo_ros_pkgs}} with \gls{ros} 1 Noetic are used as the framework for the simulation.
The underground cave environment is based on the environment presented in~\cite{koval_subterranean_2020}.
Our robot team consists of three heterogeneous robots: Clearpath Husky, Boston Dynamics Spot and a Parrot Bebop.
\begin{table}[h]
    \vspace{-10px}
    \caption{
        Overview of the capabilities of each team member.
        The check mark indicates that a robot provides an implementation for the specific capability.
        Each indicated duration shows the time the simulation director waits before reporting the result of the operation to the robot.
        For the explore capability, the reported distance in meters is the radius around the robot in which tasks are revealed on the map.
    }
    \label{tab:robot_team}
    \centering
    \begin{tabularx}{\columnwidth}{X c c c c}
        Robot Name                      & \makecell{Distance                                                                     \\ Cost \\ Factor} & Explore  & Identify           & Decontaminate  \\
        \hline
        \multirow{2}{*}{\textbf{Husky}} & \multirow{2}{*}{2.0}  & \checkmark             & \checkmark        & \checkmark        \\
                                        &                       & \textit{(60 sec, 15m)} & \textit{(60 sec)} & \textit{(8* sec)} \\
        \multirow{2}{*}{\textbf{Spot}}  & \multirow{2}{*}{1.25} & \crossmark             & \checkmark        & \crossmark        \\
                                        &                       & \textit{(- sec)}       & \textit{(2 sec)}  & \textit{(- sec)}  \\
        \multirow{2}{*}{\textbf{Bebop}} & \multirow{2}{*}{1.0}  & \checkmark             & \crossmark        & \crossmark        \\
                                        &                       & \textit{(10 sec, 30m)} & \textit{(- sec)}  & \textit{(- sec)}  \\
    \end{tabularx}
\end{table}

Table~\ref{tab:robot_team} shows the cost estimation parameters for each robot and which capabilities they provide.
All robots start within the start area, as seen in figure~\ref{fig:cave_world:navigation}, and use the same \gls{bt} consisting only of a \texttt{RemoteCapabilitySlot}.
\subsection{Dynamic Assignment}
\label{sec:evaluation:dyn_assign}
\begin{figure}[t]
    \centering
    \begin{adjustbox}{width=\columnwidth}
        \input{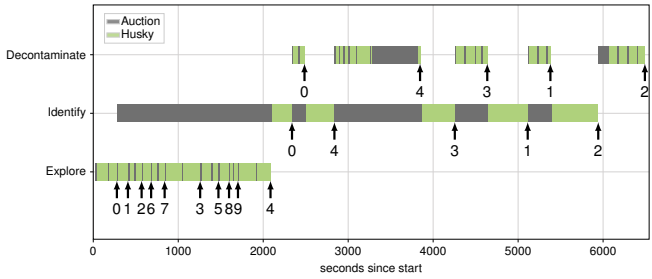}
    \end{adjustbox}
    \caption[Task allocation over time for a single Husky]{
        Task allocation during a mission run with a single Husky robot.
        Arrows denote time points when a task, with the ID as seen in figure~\ref{fig:cave_world:navigation}, was completed.
    }
    \label{fig:evaluation:husky}
    \vspace{-15px}
\end{figure}
As seen in Figure~\ref{fig:mission_bt}, the capabilities abstract robot skills as an easy-to-understand and reusable module that enables quick mission building.
During each tick of the capabilities, the best executor is evaluated, dynamically assigning the part of the mission to the most eligible robot.
Figure~\ref{fig:evaluation:husky} shows the task allocation for a mission run with a single Husky robot.
As per the robot's bidding strategy, the auction-based task assignment preferred the execution of tasks with lower overall costs.
As the distance to the task location is the major cost factor, this expresses itself in the execution of the physically closest task.
\begin{figure}[tbh!]
    \centering
    \begin{adjustbox}{width=\columnwidth}
        \input{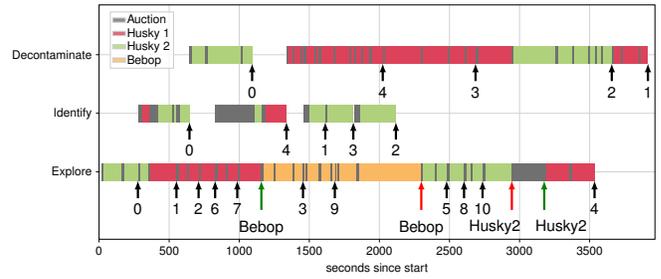}
    \end{adjustbox}
    \caption[Task allocation over time for a Husky, Spot and Bebop]{
        Task allocation over the mission runtime when using a full team consisting of two Huskys and a Bebop.
        Arrows denote time points when a task, with the ID as seen in figure~\ref{fig:cave_world:navigation}, was completed.
        Green arrows indicate a team member joining the team, whereas red arrows indicate a team member leaving.
    }
    \label{fig:evaluation:husky_bebop_spot}
    \vspace{-15px}
\end{figure}
Figure~\ref{fig:evaluation:husky_bebop_spot} depicts a mission run starting with two Husky robots.
During this run, a Bebop drone joins for a certain period, and a Husky fails for a short period.
Here, dynamic task allocation based on each robot's reported costs can be observed.
The explore task is allocated to the Bebop drone immediately after it joins the team, as it reports lower costs even though it has to travel a longer distance initially.
The existing allocation is adapted through re-auctions to minimize the overall costs and leverage the possible task parallelism from employing a robot team.

At all times, the system provided a valid, but due to the simple auction strategy not always globally optimal, assignment of tasks to robots based on their capabilities.

\section{Conclusions and Future Works}
\label{sec:conclusions}
In this paper, we have presented Behavior Tree Capabilities as an extension of the previously published \mybtlib~to conveniently structure robot skills and dynamically distribute them within a team of heterogeneous robots.
We presented the theoretical concept that comes with the notion of preconditions, IOBridges, and RemoteCapabilitySlots and how it is implemented.
The proposed system was evaluated with a simulated mission, showcasing the ability to distribute tasks in the team dynamically and to optimize the overall mission by selecting the most suitable systems.
We have shown that the Behavior Tree Capabilities are an intuitive and versatile system to abstract individual robot behavior and can be used to enable spontaneous cooperation within a team.
By using capabilities as a model in the \gls{bt} mission definition, while allowing for runtime selection of the respective robot-dependent implementations, tasks can be handled robot-specific while remaining modular.
Such flexibility greatly enhances the option of using multiple heterogeneous robots and spontaneously adding new systems to the mission.
By building the assignment system into the core library and defining an implementation mapping as a core principle, we enable an ad-hoc distribution of the capabilities to the most suitable system and, therefore, a decentralized solution to the \gls{mrtap}.
By modeling preconditions and allowing (re-)distribution, we reduce the required apriori assumptions and knowledge about a mission.
Instead the focus shifts on the development of robust, reusable skills for individual robots.
Currently, we are porting the library to ROS~2 and are considering the transfer and translation of capabilities as a form of skill learning in the future.

\newpage

\printbibliography

\end{document}